\documentclass{article}

\usepackage{arxiv}

\usepackage[utf8]{inputenc} 
\usepackage[T1]{fontenc}    
\usepackage{hyperref}       
\usepackage{url}            
\usepackage{booktabs}       
\usepackage{amsfonts}       
\usepackage{nicefrac}       
\usepackage{microtype}      
\usepackage{lipsum}
\usepackage{graphicx}
\graphicspath{ {./images/} }

\title{Naiad: Novel Agentic Intelligent Autonomous System for
Inland Water Monitoring}

\author{
  Eirini Baltzi\footnotemark[1] \\
  National Technical University of Athens \\
  \texttt{irenempal00@gmail.com} \\
   \And
  Tilemachos Moumouris\footnotemark[1] \\
  National Technical University of Athens\\
  \texttt{tilnmoumouris@gmail.com} \\
  \And
  Athena Psalta \\
  National Technical University of Athens\\
  \texttt{psaltaath@central.ntua.gr} \\
  \And
  Vasileios Tsironis \\
  National Technical University of Athens\\
  \texttt{tsironisbi@central.ntua.gr} \\
  \And
  Konstantinos Karantzalos \\
  National Technical University of Athens\\
  \texttt{karank@central.ntua.gr}\\
  }

\begin{document}

\maketitle

\begingroup
\renewcommand\thefootnote{\fnsymbol{footnote}}
\footnotetext[1]{Equal contribution.}
\endgroup

\begin{abstract}
Inland water monitoring plays a critical role in safeguarding public health and preserving ecosystems, enabling timely interventions to mitigate potential risks. Current methodologies focus on addressing specific sub-problems, such as monitoring cyanobacteria severity, chlorophyll concentration, or other water quality parameters individually. NAIAD employs an agentic AI assistant leveraging Large Language Models (LLMs) and a variety of external tools to provide a holistic solution for general-purpose inland water quality monitoring using Earth Observation (EO) data. Targeting a broad audience, from experts to individuals with no EO experience, NAIAD employs a single-prompt interface that translates user queries into actionable insights. Through Retrieval-Augmented Generation (RAG), LLM reasoning, external tool calling, computational graph execution and agentic reflection patterns, the system retrieves and synthesizes relevant information from curated knowledge sources, generating comprehensive and tailored reports. NAIAD supports a wide range of external tools, including weather or Sentinel-2 data input, remote sensing index calculation (e.g., NDCI), custom analysis tools (e.g., chlorophyll-a estimation), and established platforms (e.g. CyFi). The system’s performance is evaluated using standard metrics, such as correctness rate and output relevancy exceeding 77\% and 85\% respectively in preliminary results, against a dataset specifically created and curated for NAIAD, assessing performance against varying levels of user expertise. Preliminary results indicate promising performance and high versatility relative to the expertise level of the input prompt. Additionally, an ablation study for optimal LLM selection was conducted, with models such as Gemma 3 (27B) and Qwen2.5 (14B) demonstrating the best trade-off between computational cost and performance.
\end{abstract}

\section{Introduction}
Marine and inland water monitoring are critical components of ecosystem management, providing essential data to support sustainable environmental management \cite{ahmed2021improved,kournopoulou2024atlas}, public health \cite{bhateria2016water} and ecosystem resilience \cite{wang2024landscape, flensborg2023indicator}. Inland aquatic systems, such as lakes, rivers and reservoirs, are exceptionally vulnerable to pressures such as urbanization, intensive agriculture and pollution events \cite{li2023urbanization, saxena2025water}. For instance, toxic cyanobacteria-motivated algal blooms may contaminate drinkable water and disrupt ecosystems \cite{liu2022remote}. Thus, effective monitoring and early warning are essential, not only to support regulatory compliance and resource stewardship but also to abate risks from emerging threats like algal toxins, eutrophication, and ecosystem collapse. In-situ monitoring networks and modern IoT-enabled sensors offer high-frequency and localized measurements of key aquatic parameters, providing invaluable ground-truth data for real-time assessment and regulatory compliance \cite{garrido2022smart}. However, these solutions are inherently limited in spatial coverage and scalability, especially across large or remote water bodies.

Conventionally, remote sensing techniques based on artifical intelligence (AI) have been deployed in monitoring of individual water quality pointers. For instance, deep learning models can accurately detect phenomena such as chlorophyll concentrations \cite{wang2022estimation}, debris \cite{kikaki2020remotely}, turbidity levels \cite{hossain2021remote}, oil spills \cite{kikaki2024detecting} and cyanobacterial blooms \cite{dorne2024cyanobacteria} from satellite imagery. 
While these approaches are successful in specific targeted Earth Observation (EO) applications, they typically require user-driven manual integration of results from multiple analyses to assemble a comprehensive and coherent picture of overall water quality.
Moreover, these conventional workflows often lack interoperability and automation, imposing a substantial technical burden on non-expert users, such as environmental practitioners, who must reconcile diverse, heterogeneous outputs from disparate tools. Consequently, many existing AI-based monitoring systems have yet to fully address these challenges: they frequently fall short in seamlessly integrating heterogeneous data sources, providing natural language interfaces that accommodate users with varying expertise levels, and dynamically orchestrating multiple analytical tools in response to evolving user queries specifically for water monitoring.

To overcome these limitations, 
we introduce NAIAD, a novel agentic AI assistant 
designed to provide integrated, accessible and adaptive inland water monitoring by unifying heterogeneous data sources, harnessing advanced Large Language Models (LLMs) and orchestrating diverse analytical tools including satellite imagery, in situ sensor networks, and meteorological data. Through a natural language, single-prompt interface, NAIAD's modular design architecture translates user queries through dynamic orchestration to retrieve comprehensive, multi-parameter assessments of inland water quality.
We leverage retrieval-augmented generation (RAG) with LLM reasoning to reformulate inputs, retrieve relevant domain-specific documents and select the most appropriate data sources and external tools in real-time. NAIAD constructs and executes computational workflows on the fly, enabling complex multi-step analyses such as downloading satellite images, calculating water quality indices (e.g., NDCI), running predictive models and aggregating meteorological data for contextual interpretation. A key innovation of NAIAD is its on-the-fly construction of Directed Acyclic Graphs (DAGs) that represent the sequence of analytical operations constituting the workflow. Each node corresponds to a discrete task—such as satellite image retrieval, index computation (e.g., NDCI), or report generation—with edges encoding data dependencies and execution order. This graph-based orchestration allows NAIAD to flexibly handle complex queries requiring multi-step analyses while maintaining transparency and traceability of the process. Additionally, NAIAD’s design incorporates agentic reflection mechanisms for output review, optimizing report accuracy and relevance by integrating diverse EO-derived metrics and summaries. Specifically, the contributions of this work are:
\begin{itemize}
\item We design and implement NAIAD, an agentic AI assistant for inland water quality monitoring; code is publicly available at \hyperlink{code}{https://github.com/rslab-ntua/naiad}.
\item NAIAD integrates a wide range of diverse external tools including Sentinel-2 imagery, meteorological data, water quality indices and platforms like CyFi for automated, multi-parameter assessment.
\item Extensive tests on realistic user queries showcase the robust performance of NAIAD across expertise levels, with more than 77\% correct tool invocation and more than 85\% output relevance.
\item Ablation studies confirm Gemma 3 (27B) and Qwen 2.5 (14B) as optimal open-source LLMs for efficient, high-quality operation.
\end{itemize}

\begin{figure}[t]
	\centering
	\includegraphics[width=12cm]{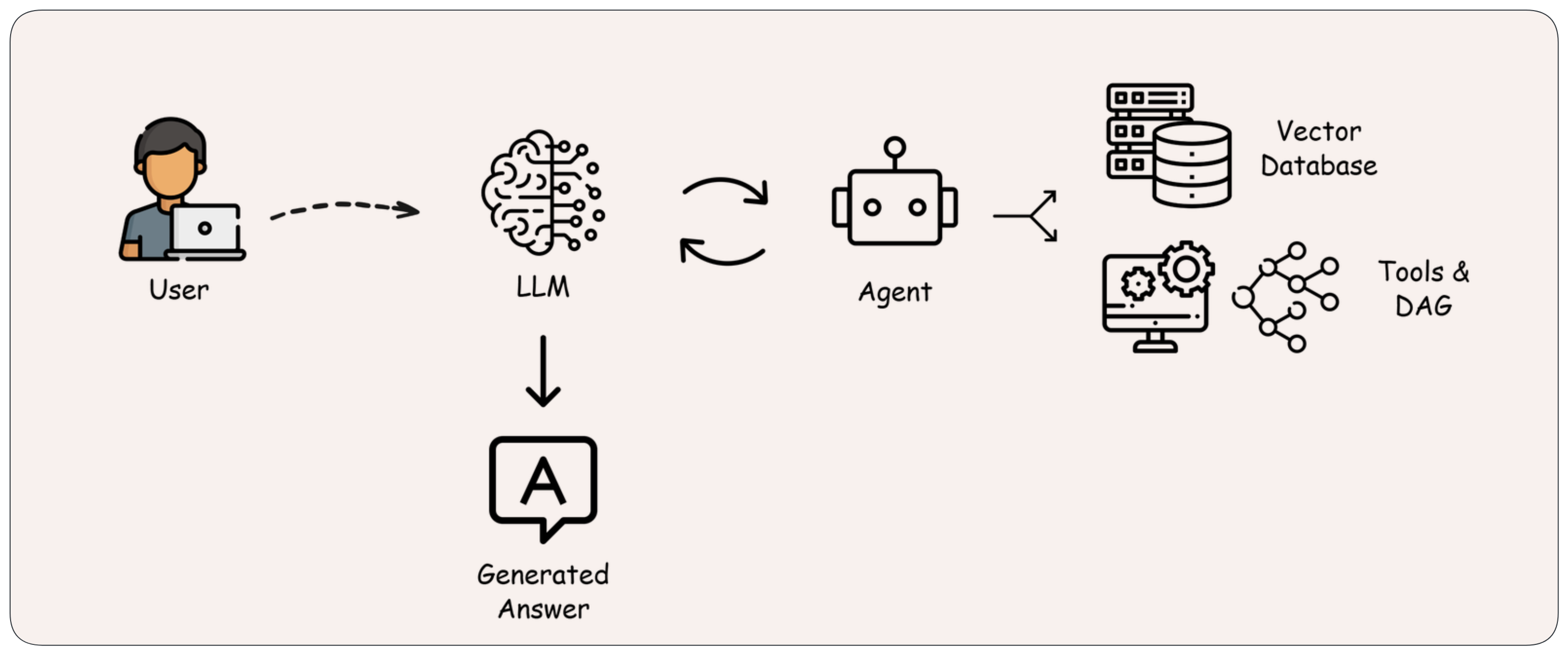}
	\caption{Graphical abstract of the NAIAD agentic AI system for inland water monitoring. User queries are processed by a LLM, which collaborates with an autonomous agent to dynamically retrieve information from a vector database and orchestrate external analytical tools. The workflow enables the generation of comprehensive, user-adapted water quality assessments via a simple prompt interface.}
	\label{fig:graph_naiad}
\end{figure}

\section{Related Work}
Inland water quality monitoring has progressed from traditional remote sensing methods to modern AI-enhanced EO workflows, reflecting a dynamic and interdisciplinary research landscape.
Traditional EO approaches have primarily relied on single-purpose algorithms for estimating parameters such as chlorophyll-a, turbidity or cyanobacteria concentrations from sensors like Sentinel-2 MSI and Landsat-8. Spectral indices of Normalized Difference Chlorophyll Index (NDCI) have proved particularly appropriate for eutrophication assessment in inland waters \cite{rs11010064, rs13051043,su16052090}. Classical machine learning models, while impactful on the validation of band-ratio algorithms and index-based schemes for national-scale surveys, frequently relied on manual feature extraction and lacked integration into end-to-end, automated workflows at scale.

In recent years, research has increasingly explored how LLMs and agentic AI approaches can be integrated into geospatial and Earth Observation (EO) workflows. This shift seeks to lower barriers to entry by enabling intuitive natural language interaction with complex analytical processes, thereby making EO data more accessible to a wider range of users. 
A central trend in this domain is the application of Retrieval-Augmented Generation (RAG), where LLMs dynamically access external domain knowledge to improve the relevance and ensuring the accuracy of generated responses. For instance,  \cite{wen2025rs} introduced RS-RAG, which links high-resolution satellite imagery with a multimodal knowledge base for enhanced captioning and classification, while  \cite{yu2025spatialrag} developed Spatial-RAG to combine semantic with geospatial filtering for question answering over maps. Building on these foundations, researchers have begun translating agentic AI concepts into real-world EO applications. RS-Agent \cite{xu2025rs} is a modular assistant for remote sensing analysis, featuring structured toolkits, expert task templates, and a dual-RAG engine for dynamic retrieval. UnivEARTH \cite{kao2025univearth} is a benchmark for evaluating Earth science agents on practical analytical questions. Their findings underscore the need for fine-tuned models and robust tool integration, as many state-of-the-art LLMs still struggle with EO-specific code generation and reliable execution.\

To address the challenge of task scalability and workflow complexity, recent research has expanded from single-task agents to multi-agent orchestration frameworks. Complementing these efforts, the GeoLLM-Engine framework \cite{geollm_engine} provides a scalable environment for evaluating agentic EO assistants on complex geospatial tasks. It simulates long-horizon, multi-step queries using dynamic APIs and UI elements, supporting functional evaluation across over 500,000 prompts and 1.1 million satellite images. Futhermore, GeoLLM-Squad \cite{geollm_squad} introduced a multi-agent orchestration approach that separates agentic planning from geospatial task execution. Built atop GeoLLM-Engine\cite{geollm_engine} and AutoGen\cite{wu2024autogen} , it enables modular integration of EO sub-agents (e.g., for forestry, climate, agriculture). Its hybrid orchestration method improves upon both ledger-based and composition-based systems, achieving a 17\% increase in agentic correctness over single-agent baselines. The system supports over 500 API tools and scales robustly across tasks and LLM architectures, including smaller open-source models. 

Further extending the RAG paradigm, \cite{ZHANG2025104312} introduced GTChain, a domain-specific LLM trained to autonomously plan and execute toolchains for geospatial tasks. This system learns to generate multi-step tool-use sequences from simulated instruction data and significantly improves planning reliability over baseline LLMs. Similarly, \cite{CHEN2025328} proposed Geo-MMRAG, a multi-modal RAG framework for lithological interpretation in remote sensing. Their system enhances foundation model outputs by retrieving expert geological knowledge (from books, reports, and datasets) to form context-aware prompts. Geo-MMRAG demonstrates how domain-specific knowledge integration can improve interpretation of ambiguous and fragmented EO imagery without additional fine-tuning.

While these recent agentic and retrieval-augmented methods mark significant progress, a gap remains for practical, domain-tailored solutions capable of user-adaptive orchestration in monitoring workflows. To address this,
NAIAD offers a domain-specialized prompt-driven assistant for inland water monitoring. 
Rather than employing a multi-agent architecture as in frameworks like multi-agent frameworks\cite{geollm_squad}, NAIAD adopts a single-agent approach that dynamically constructs analytical workflows as directed acyclic graphs (DAGs) on-the-fly for seamless tool orchestration.
It also distinguishes itself from GTChain \cite{ZHANG2025104312} by focusing on live execution of user queries rather than simulated toolchain learning and by integrating
tools for Sentinel-2 imagery, NDCI, weather data, and CyFi analysis. 
Through this design, NAIAD delivers a deployable, user-friendly system that bridges advanced EO analytics with natural language accessibility, effectively translating state-of-the-art AI capabilities into actionable, real-world inland water monitoring support.

\section{Methodology}

\subsection{Overall Design}

\begin{figure}[t]
	\centering
	\includegraphics[width=\textwidth]{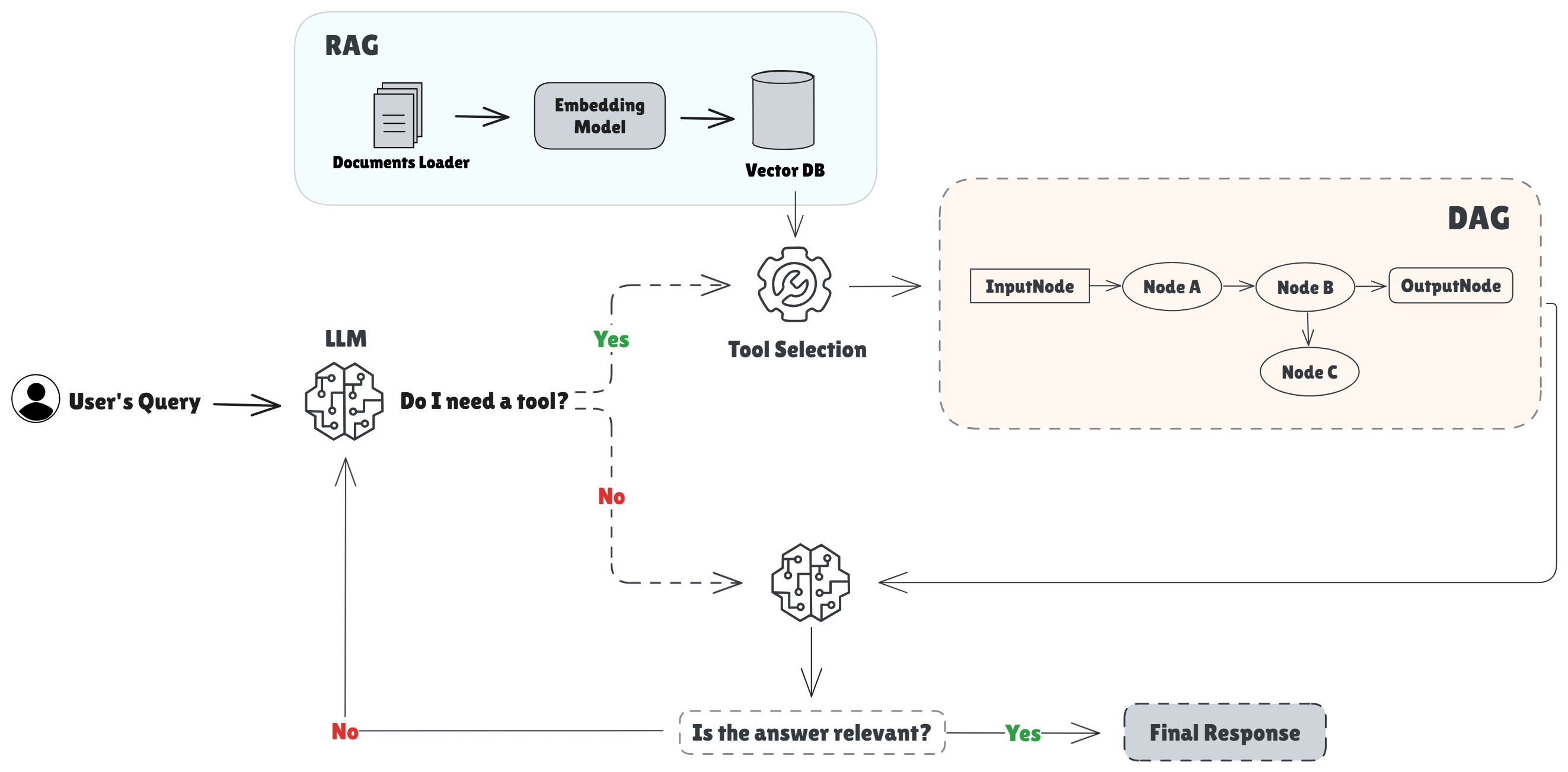}
	\caption{\textbf{Overview of the NAIAD system architecture and workflow.} After the user submits a query, NAIAD interprets this query using RAG to retrieve relevant domain knowledge and add context. The LLM then determines whether external tools are needed, selecting and orchestrating them by dynamically constructing a Directed Acyclic Graph (DAG) that encodes the analytical workflow. Throughout the process, reflection mechanisms ensure output relevance and accuracy. The final report synthesizes all findings and is tailored to the user’s expertise.}
	\label{fig:naiad}
\end{figure}

NAIAD relies on a simple yet efficient architecture that takes advantage of tools to monitor key environmental variables such as Chlorophyll and temperature to effectively help decision makers and researchers gather information about inland water bodies in a short time. Figure \ref{fig:naiad} describes the high level design of our system. The path of the input query can be divided in two parts that contribute to the final output.

The process begins at the user interface, where a prompt-driven interaction allows users to submit queries in natural language, simplifying the report generation workflow. The input query is rewritten by the LLM to better align with the specific task of inland water monitoring; this rewriting step improves clarity and relevance. The system retains the original query alongside the rewritten version for downstream processes. To enhance the semantic understanding of the query, we integrate a Retrieval-Augmented Generation (RAG) engine. The RAG module, powered by the \textit{BAAI/bge-large-en-v1.5} embedding model and managed through the Llama Index Framework, retrieves top-10 relevant documents from specialized content tanks. These documents are selected based on similarity and are orchestrated by the LLM to ensure the correct information is injected at the appropriate stages of processing. This enrichment mechanism extends even to tools that traditionally would not rely on context-based content, effectively making NAIAD’s RAG strategy system-wide.

 One of the key architectural features of NAIAD is its ability to dynamically construct Directed Acyclic Graphs (DAGs) that define task-specific workflows. The DAG is instantiated at runtime by the LLM and encodes a structured execution plan composed of modular operations. Each node in the DAG represents a discrete functional unit—such as a retrieval action (e.g., NDCI fetching), a transformation (e.g., chlorophyll estimation), or a report generation module. Nodes are labeled with their operation type, required inputs, and any downstream dependencies. Edges between nodes define strict data and execution dependencies. An edge from node A to node B means that the output of A must be available before B executes. These edges are not hard-coded, but inferred by the LLM using internal prompting strategies that analyze tool metadata, user intent, and I/O compatibility. For example, if a chlorophyll estimation tool requires an NDCI value, and another tool provides it, an edge is automatically formed. Tools are described via structured metadata including I/O schemas, temporal scope, and valid invocation contexts and this metadata is embedded into the system prompts provided to the LLM. The DAG is first constructed as a declarative structure in memory (e.g., a JSON graph object), then validated before execution. Validation ensures that all nodes have their input dependencies satisfied, there are no cycles, and that tool sequences follow permissible patterns (e.g., a report node must be terminal). NAIAD supports conditional branches in the DAG (e.g., fallback tools or alternative pipelines) and can dynamically skip parts of the workflow if certain results are already available or if external services are temporarily unavailable. This graph-based execution plan enables precise, modular orchestration of heterogeneous tools, accommodates failures via node-level retries, and simplifies debugging and interpretability of agent behavior. Additionally, the DAG design allows the system to simulate or preview execution paths before actual tool invocations, further enhancing safety in high-stakes decision support scenarios.

Integrated tools within this workflow include empirical Chlorophyll models, NDCI mosaics, and external data providers such as weather and CyFi services. These modules enable users to monitor key water quality indicators and support timely intervention. The system also facilitates large-scale environmental observation through automated Sentinel-2 data queries, enabling satellite-driven assessments. Additionally, we developed custom external endpoints to manage system-triggered operations such as vegetation index calculations or time series generation. These services are exposed through a lightweight web server architecture, enabling rapid deployment and future extensibility.

Following the execution of the workflow, NAIAD applies a reflection and correction mechanism to evaluate the outputs. This post-processing phase allows the system to revise and refine the results before delivering the final report. A log of irrelevant or incorrect responses is maintained and fed back into the system, providing a form of self-correction and progressive improvement over time. This capability strengthens the robustness of the system, especially in iterative use cases.

To support flexible deployment across varied operational contexts—including governmental applications where data privacy is paramount—NAIAD is designed for self-hosted execution. For LLM inference, we employ the Ollama framework, which facilitates efficient local deployment and modular management of open-source language models. This infrastructure choice allows us to interchange LLM backbones during ablation studies while maintaining full control over security and compute resources.

\subsection{External tools integration}
Our system is designed to enable users to easily add new tools with minimal effort, ensuring both scalability and granular control over feature integration. These tools can be hosted either within the system or externally, communicating via requests to dedicated servers which is particularly useful when dealing with frequently updated data from on-site observations. NAIAD integrates a diverse and extensible set of external tools designed to provide comprehensive and rapid assessment capabilities for inland water monitoring. These tools are modular and can be deployed locally or as external services, allowing seamless integration and future scalability. Supported tools that were selected for the initial use and evaluation in our system are the following:

\noindent \textbf{Satellite data acquisition tool.} This tool automates the retrieval of Sentinel-2 imagery for specific water bodies and time intervals, as determined by the user’s query. The LLM processes the query to extract spatial (coordinates or water body name) and temporal (start/stop date) parameters, after which the system requests the appropriate satellite tiles containing the target region. This modular design allows integration of additional satellite sources or data providers, such as Landsat  or MODIS, to meet specific geographic or temporal requirements.

\noindent \textbf{Remote sensing index calculation tool.} This tool is responsible for computing remote sensing indices, such as the Normalized Difference Chlorophyll Index (NDCI) or Normalized Difference Water Index (NDWI), to provide robust proxies for primary water quality indicators. Utilizing the downloaded satellite imagery, the tool performs zonal statistics across designated regions such as lakes specified by the user. A typical operation is the calculation of maximum or average NDCI values within a vector-defined area. The system’s flexible architecture supports the inclusion of new or custom indices as monitoring needs evolve.

\noindent \textbf{Water quality parameter estimation and prediction tool.} This integrated toolset encompasses empirical estimation and predictive analytics crucial for comprehensive inland water monitoring. Central to this is the \textit{chlorophyll calculator}, which leverages remotely sensed indices such as NDCI to estimate chlorophyll-a concentrations which is a key indicator of biological productivity and water quality status. Acting as a core node within the computational workflow, this tool can provide standalone parameter estimates or feed results into subsequent analysis and reporting stages. Complementing this, the system incorporates advanced predictive capabilities through integration with the CyFi platform \cite{dorne2024cyanobacteria}, which specializes in forecasting harmful cyanobacterial bloom intensities based on spatiotemporal input parameters. This predictive modeling is pivotal for public health and ecosystem management due to the potential toxicity and ecological impact of cyanobacteria. The LLM dynamically extracts location coordinates, time frames, and other query details to construct precise requests for bloom level estimation. Upon receiving data from CyFi, the system enhances interpretability by augmenting raw outputs with domain knowledge using a dedicated retrieval-augmented generation engine, thereby producing context-rich, actionable insights. This comprehensive approach ensures that NAIAD supports both parameter quantification and risk forecasting within a unified, agentically orchestrated monitoring framework.

\noindent \textbf{Environmental context enrichment tool.} To provide a more comprehensive interpretation of water quality data, this tool retrieves relevant meteorological parameters (such as temperature, wind speed, and precipitation) for the water body and timeframe in question. The LLM extracts location information, which is then used by the tool to query external weather APIs. This contextual data is critical for identifying environmental drivers and for interpreting water quality fluctuations. The enrichment strategy can also incorporate forecast trends, supporting short-term prediction and adaptive management actions.

\noindent \textbf{Automated report generation tool.} Report generation is a core capability of NAIAD, synthesizing all results from the executed workflow—including raw data, derived indices, predictive outputs, and contextual enrichments into a coherent and user-tailored report. The report tool accepts as input a reformulated version of the original user query or the aggregated knowledge from previous steps. Leveraging a dedicated RAG engine, this subsystem ensures that outputs are coherent, relevant and adapted in language and detail to the user’s background. The modular report generator allows for straightforward updates or inclusion of new analytical insights, keeping pace with scientific and operational advancements.

\subsection{Evaluation Strategy}\label{sec:3.3}

We adopt the task generation and evaluation protocol proposed in GeoLLM-Squad~\cite{geollm_squad}. Specifically, we provide user-task “seeds” to GPT-4o and generate a set of representative prompts that simulate realistic inland water monitoring workflows. Each prompt is manually annotated with the expected steps (i.e., tool invocations and content flow) required for successful execution. These curated solutions are then used to guide the agent and form a pseudo ground-truth for correctness evaluation. Our evaluation focuses on measuring \textit{agentic correctness}, specifically following the metric of correctness rate. For each task, we assess: (i) whether the agent interpreted the input correctly (input correctness), (ii) whether the correct tools were invoked (correct tool calling), and (iii) whether the tool execution followed the expected order (correct order). The final \textit{correctness rate} is computed as the average over these three metrics. We also recorded whether the final output was relevant (relevancy) and noted any anomalies or failures (e.g., external service downtime or redundant tool usage). A key focus of the evaluation was the structure and correctness of the Directed Acyclic Graphs (DAGs) generated for each task, as these represent the backbone of the agentic workflows. For each test prompt, we manually inspected the DAG to verify whether it contained the appropriate nodes (i.e., required tools), correct edge directions (i.e., logical execution order), and an overall structure that matched the intended logic of the user query.

\section{Results}

\subsection{Study Area}
\begin{figure}[ht]
    \begin{center}
    \begin{tabular}{ccc}
    \includegraphics[height=3cm]{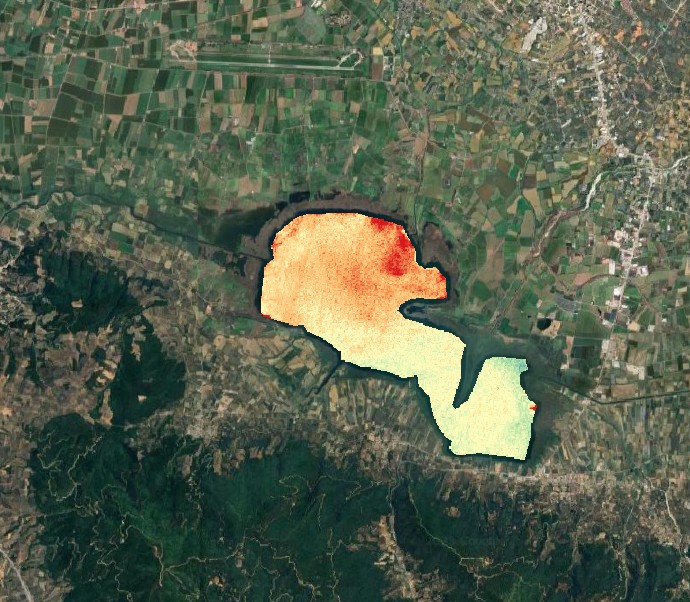} &
    \includegraphics[height=3cm]{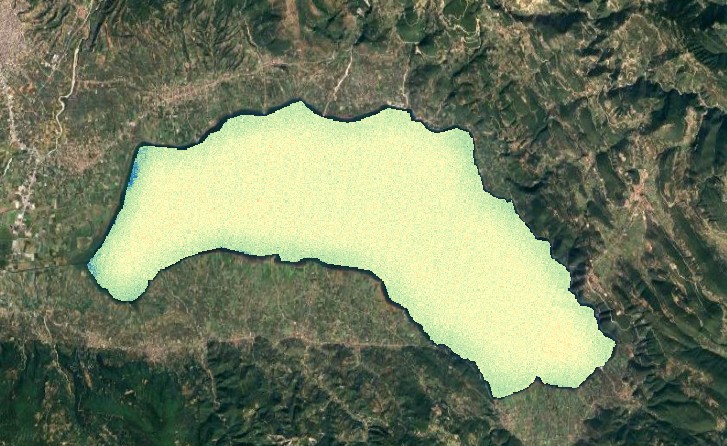} &
    \includegraphics[height=3cm]{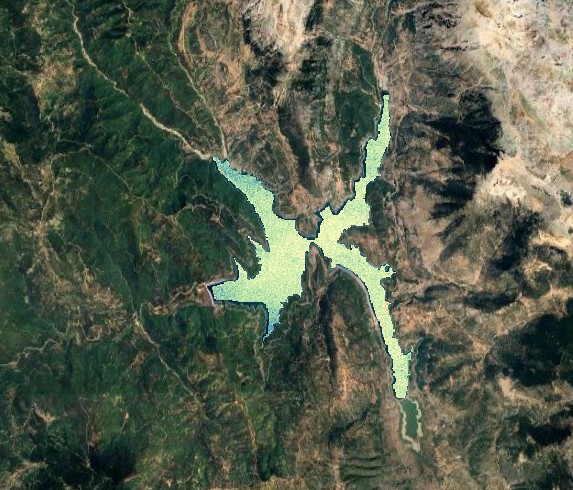} \\
    (a) & (b) & (c) \\
    \end{tabular}
    \end{center}
    \caption{ROIs of our study with NDCI calculated overlay, (a) Lake Lysimachia, (b) Lake Trichonida and (c) artificial Lake Mornos.}  
    \label{fig:study_area}
\end{figure}

Our study was focused on three main water bodies in Central Greece, lakes Lysimachia, Trichonida and Mornos (Figure \ref{fig:study_area}) to showcase the capabilities of the system. The selection of these lakes was based upon the observed eutrophic state and also their water contribution to the local or wider area.

\noindent Lake Lysimachia is a freshwater lake located in the regional unit of Aetolia-Acarnania and its catchment basin covers an area of 246~km$^2$. The lake has a surface area of 13~km$^2$ with relatively shallow waters, a maximum depth of 9~m and an elevation of 14~m above sea level. Appart from being fed by several streams it receives surplus water from Lake Trichonida, with which it is connected via a 2.8~km long channel. This water body is classified as eutrophic. It receives urban wastewater from major nearby city and surrounding settlements and is further impacted by agricultural and livestock activities in the region. Significant areas of lakeshore have been encroached upon and converted to agricultural land.

\noindent Lake Trichonida is a freshwater lake located in the regional unit as  Lysimachia and lakes' catchment basin covers an area of 421~km$^2$. With a surface area of 97~km$^2$, Lake Trichonida is the largest lake in Greece. The maximum depth of 58~m and has an elevation of 16~m above sea level. The lake is fed by surface runoff from its hydrological basin, as well as by both surface and underground springs. Despite pressures from agricultural and livestock activities and urban wastewater, the lake maintains its natural and water resources. This resilience is due to the rapid renewal of its waters, which are significant in volume, highly transparent, well-oxygenated, and have low nutrient concentrations. As a result, Lake Trichonida is classified as oligotrophic.

\noindent Finally, lake Mornos is an artificial reservoir which is located in relative close proximity to lake Trichonis and was originally constructed to meet the increasing water supply needs of Athens. Created in 1979 after the completion of the Mornos Dam on the Mornos river, the lake has a surface area of approximately 15.5~km$^2$ and is situated at an elevation of 446~m, making it the ninth largest among all artificial lakes in Greece. The Mornos aqueduct, one of the largest in Europe with a total length of 188~km, transports water from the lake to the greater Athens area, supporting both urban and agricultural needs.

\subsection{Underlying LLMs}
In our agentic workflow, we leverage different LLMs in order to find the optimal trade-off regarding performance and computational cost. The first model utilized was \textbf{Qwen2.5}, a state-of-the-art LLM series that has demonstrated strong performance across a range of language understanding and generation tasks. \textbf{Qwen2.5 (14B)}, as described in its technical report \cite{qwen2025qwen25technicalreport} , benefits from extensive pre-training on high-quality datasets and advanced post-training techniques, making it a robust choice for foundational language processing within our agentic workflow. Furthermore, we experimented with the integration of \textbf{Gemma3 (27B)} \cite{gemmateam2025gemma3technicalreport} , an open model family known for its extended context windows up to 128K tokens.

\subsection{Evaluation Results}
To rigorously assess the performance of NAIAD, we conducted a comprehensive evaluation using an in-house developed gold-standard dataset, as described in Section \ref{sec:3.3}. Our evaluation focused on two primary metrics: the \textbf{relevancy} and \textbf{correctness rate} of the system’s final outputs, as well as the accuracy of tool usage within the agentic workflow. All experiments were run using an NVIDIA RTX A6000 48Gb graphics card. LLMs were served using Ollama, through a dedicated container ensuring rapid development with different models.

\noindent The gold-standard dataset comprised a diverse set of user queries, simulating realistic inland water monitoring scenarios across varying levels of user expertise. Each query was manually annotated with the expected sequence of tool invocations and the logical flow of information, serving as a reference for correctness assessment. For each test case, we evaluated whether NAIAD correctly interpreted the input, invoked the appropriate tools in the correct order, and generated outputs that were both relevant and actionable. In Table \ref{tab:model_results} we present the final results after the evaluation phase. 

\begin{table}[ht]
\caption{Final evaluation results for models used in our system.}
\label{tab:model_results}
\begin{center}
\begin{tabular}{|c|c|c|c|}
\hline
\rule[-1ex]{0pt}{3.5ex} Model & Number of Parameters & Correctness \% & Relevancy \%\\ 
\hline
\rule[-1ex]{0pt}{3.5ex} Qwen2.5  & 14B & 82.98 & 78.72 \\ 
\hline
\rule[-1ex]{0pt}{3.5ex} Gemma3  & 27B & 82.98 & 68.09 \\ 
\hline

\end{tabular}
\end{center}
\end{table}

\noindent\textbf{Qwen2.5(14B):}
Qwen2.5, a state-of-the-art 14B parameter LLM, demonstrated robust performance across all evaluation criteria. The model achieved a correctness rate of 82.98\%, indicating a high degree of alignment between the generated Directed Acyclic Graphs (DAGs) and the annotated ground-truth workflows. In terms of output relevancy, Qwen2.5 scored 78.72\%, reflecting its ability to generate responses that were not only accurate but also contextually meaningful and actionable for end users.

\begin{figure*}
\scriptsize
\centering
\setlength{\tabcolsep}{1.2pt}
\newcommand{\sz}{17cm}
\begin{tabular}{c}

\includegraphics[width=\sz,keepaspectratio]{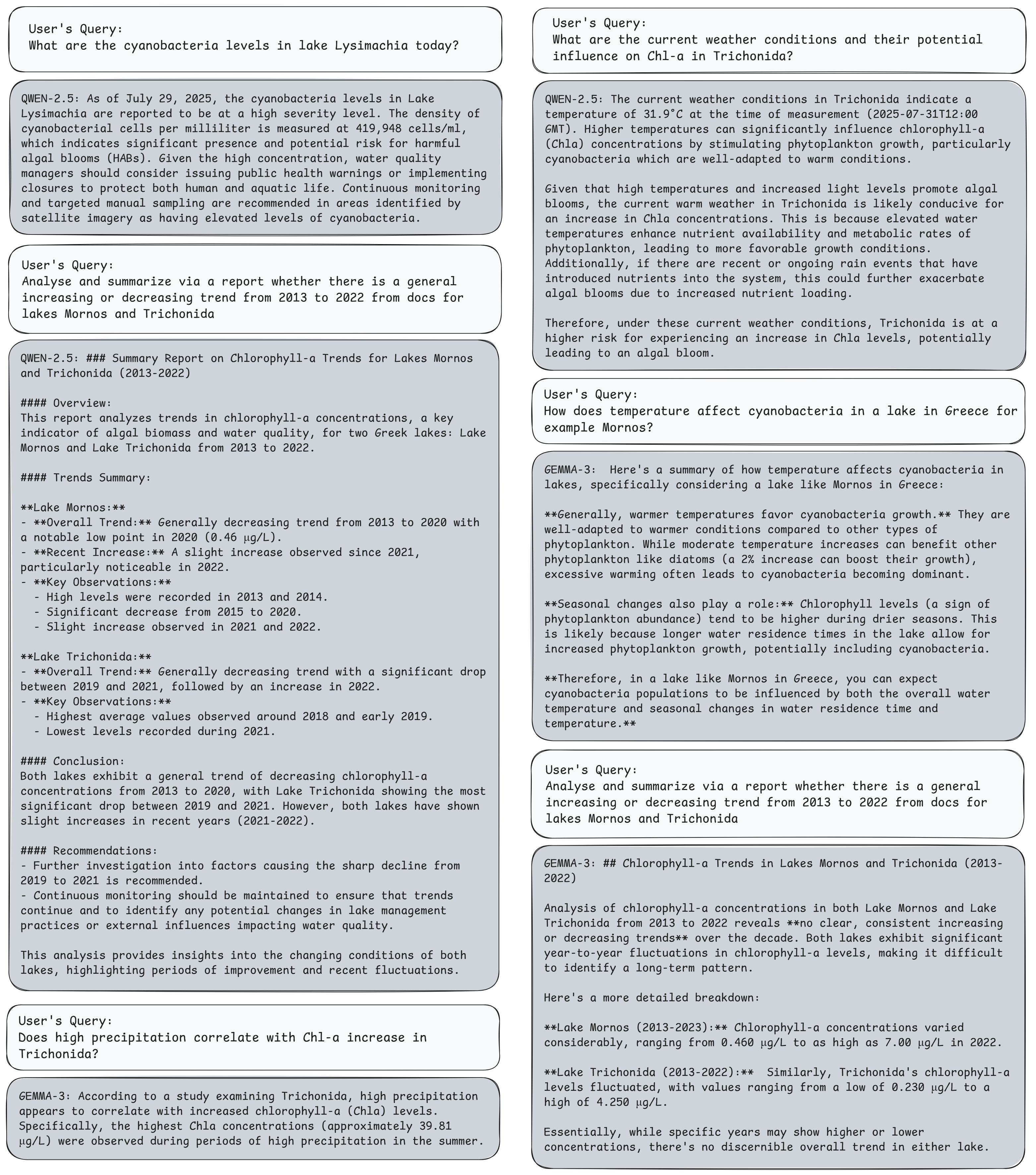}

\end{tabular}
\caption{Example user queries and corresponding LLM-generated responses from NAIAD, across three Greek lakes (Mornos, Trichonida, Lysimachia). The queries address topics such as chlorophyll-a trends, precipitation and temperature effects, cyanobacteria levels, and weather influences. Responses are generated by models Qwen-2.5 and Gemma-3, showcasing NAIAD's ability to retrieve, reason, and summarize scientific insights based on structured document sources and environmental data.}
\label{fig:queries}
\end{figure*}

\noindent\textbf{Gemma3 (27B):}\
Gemma3, a 27B parameter open LLM with extended context capabilities, matched Qwen2.5 in correctness, also achieving 82.98\%. However, its relevancy score was slightly lower at 68.09\%. This suggests that while Gemma3 is equally capable in orchestrating the correct analytical workflow, its generated outputs were, on average, less tailored or directly relevant to the user queries compared to Qwen2.5.

\noindent These results highlight that both models are highly effective in managing complex tool orchestration and workflow execution within NAIAD. However, Qwen2.5 offers a notable advantage in generating more relevant and user-adapted outputs, making it particularly suitable for scenarios where actionable insights and clarity are critical. The comparable correctness rates also indicate that the agentic workflow design is robust to the choice of LLM, provided the model is sufficiently capable.

\noindent In Figure \ref{fig:queries} we provide examples of queries and system responses during the evaluation process. The system's ability to construct and execute a DAG that contains the appropriate tools is visible through the flow of the queries and the needed data to answer. In more demanding workflows, the system was able to bring together external sources such as data derived from machine learning methods (CyFi) as well as weather or locally stored data which was retrieved through the RAG Engine utilized in the system. 

\noindent Overall, the evaluation confirms that NAIAD, powered by either Qwen2.5 or Gemma3, delivers reliable and context-aware inland water monitoring support. The system’s ability to maintain high correctness and relevancy across diverse queries and user profiles underscores its potential for real-world deployment in environmental management and decision support.

\section{Future Work}
NAIAD presents a promising agentic architecture for inland water monitoring, yet several extensions and enhancements are envisioned to further broaden its applicability and robustness. 

\noindent\textbf{Geographical Expansion and Domain Generalization:} Future versions of NAIAD will target broader geographical deployments beyond the three Greek lakes currently studied. This inlcudes scaling the system to new inland water bodies across Greece, Europe and globally, enabling comparative water quality assessments and cross-region trend analyses. Supporting diverse ecological and climatic zones will require fine-tuning tool parameters, adapting regional data sources (e.g., local hydrological models), and generalizing the underlying LLM prompts to accommodate the location-specific semantics. 

\noindent\textbf{Toolset Enrichment and Integration of New Modalities:} we plan to incorporate additional tools such as turbidity and trophic state index estimators using multi-sensor fusion, and support for Landsat or MODIS imagery to enable historical and coarse-scale observations. Integration with real-time in-situ sensor data will strengthen validation capabilities, while nutrient load modeling (e.g., nitrogen and phosphorus from land use patterns) and shoreline or land cover change detection tools will provide broader ecosystem context.

\noindent\textbf{Multi-Agent Collaboration:} Lastly, while NAIAD currently adopts a single-agent paradigm, future designs may explore multi-agent collaboration frameworks, allowing separate agents to handle distinct domains (e.g., climate, marine, biodiversity) and exchange intermediate results.

\section{Conclusions}
In this work, we presented NAIAD, a novel agentic intelligent assistant designed for holistic inland water quality monitoring through natural language interaction, dynamic tool orchestration, and real-time data integration. By leveraging LLMs, Retrieval-Augmented Generation, and a flexible Directed Acyclic Graph-based execution engine, NAIAD translates user queries into actionable insights using diverse tools ranging from EO data retrieval to predictive modeling. Evaluation results across multiple lakes and realistic scenarios demonstrate its high correctness and contextual relevance, particularly when powered by open models like Qwen2.5. The system's modularity, extensibility, and user-adaptiveness make it a promising solution for operational deployment in environmental decision support workflows. Future work will focus on expanding toolsets, geographic coverage, and system intelligence, positioning NAIAD as a foundation for general-purpose, agentic environmental monitoring systems.

\section*{Acknowledgments}    
 
This work was supported by the research project BiCUBES “Analysis-Ready Geospatial Big Data Cubes and Cloud-based Analytics for Monitoring Efficiently our Land \& Water” funded by the
Hellenic Foundation for Research and Innovation (HFRI) (grant: 03943)

\end{document}